\documentclass{article}
\usepackage{spconf,amsmath,graphicx}
\usepackage{multirow}
\usepackage{multicol}
\usepackage{booktabs}
\usepackage{todonotes}
\usepackage{mathtools}
\usepackage{xcolor} 
\usepackage[noadjust]{cite}
\usepackage[latin1,utf8]{inputenc}
\usepackage[T1]{fontenc}
\usepackage{hyperref}
\usepackage{caption}
\usepackage{subcaption}
\usepackage{booktabs}
\usepackage{multirow}

{}

\DeclareMathOperator*{\argmin}{arg\,min}

\allowdisplaybreaks[2]
\title{Explainers in the Wild: Making Surrogate Explainers Robust to Distortions through Perception}
\name{Alexander Hepburn\thanks{This work was partially funded by EPSRC grant EP/N509619/1 and by the UKRI Turing AI Fellowship EP/V024817/1.} 
\quad Raul Santos-Rodriguez}
\address{ Department of Engineering Mathematics, University of Bristol, UK}
\begin{document}
%
\maketitle
\begin{abstract}
Explaining the decisions of models is becoming pervasive in the image processing domain, whether it is by using post-hoc methods or by creating inherently interpretable models. While the widespread use of surrogate explainers is a welcome addition to inspect and understand black-box models, assessing the robustness and reliability of the explanations is key for their success. Additionally, whilst existing work in the explainability field proposes various strategies to address this problem, the challenges of working with data in the wild is often overlooked. For instance, in image classification, distortions to images can not only affect the predictions assigned by the model, but also the explanation. Given a clean and a distorted version of an image, even if the prediction probabilities are similar, the explanation may still be different. In this paper we propose a methodology to evaluate the effect of distortions in explanations by embedding perceptual distances that tailor the neighbourhoods used to training surrogate explainers.  We also show that by operating in this way, we can make the explanations more robust to distortions. We generate explanations for images in the Imagenet-C dataset and demonstrate how using a perceptual distances in the surrogate explainer creates more coherent explanations for the distorted and reference images.
\end{abstract}
\section{Introduction}\label{sec:intro}
The state-of-the-art methods in image classification almost exclusively rely on black-box deep neural networks. Whilst it has been argued that inherently interpretable models should be the focus of research~\cite{rudin2019stop}, transparency of models is also achieved via post-hoc methods~\cite{chen2018looks, brendel2019approximating, rafael2020}, where a more interpretable model is fit to the output of the black-box model. One of the main post-hoc explainability tools are surrogate explainers~\cite{ribeiro2016should}, where a simple but interpretable model is trained in the local neighbourhood of a query point with the objective of approximating the decision boundary of the black-box model. 

\begin{figure}[h]
    \centering
    \includegraphics[width=.8\columnwidth]{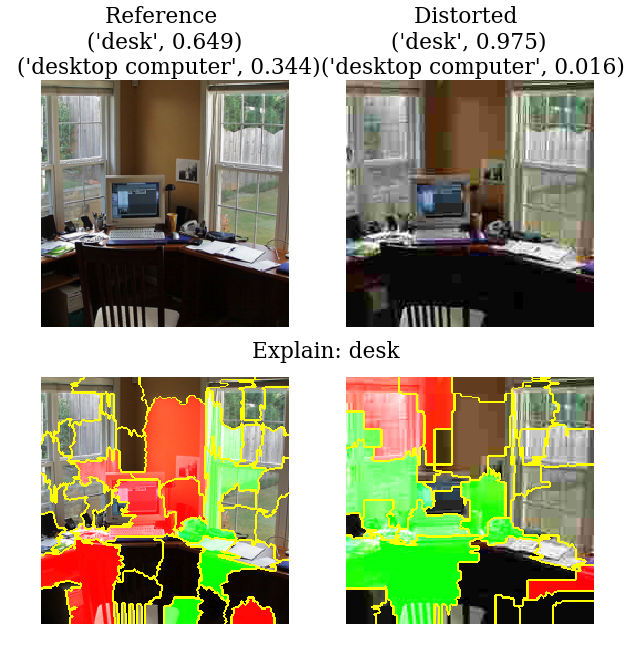}
    \caption{Explanations for a reference image and an image that has been subject to JPEG compression. The yellow lines define the superpixels found by the explainer. The green/red sections denote superpixels that have a positive/negative influence on the class.}
    \label{fig:example}
\end{figure}


Whilst more and more black-box models are being deployed in practice, there is an increasing need to be able to explain the decisions made by a model in real-world scenarios that go beyond those of carefully curated image datasets. For instance, a common situation is for these models to be presented with images that are of worse quality than the images used to train and validate the model. Images can be subject to distortions or perturbations for a number of reasons, whether it is the weather distorting the view of the subject in the image, or compression artefacts from attempting to save the image as a smaller file size. Although the performance of the models decline in these contexts~\cite{hendrycks2019benchmarking}, even if the network is able to correctly predict the class of the image, there are no guarantees on the explanation remaining the same. Our first contribution is to assess the stability and robustness of the explanations to make sure that these are not driven by undesired factors such as distortions. To illustrate our goal, in Fig.~(\ref{fig:example}), we show that even situations where the prediction remains correct for the top-2 classes of a distorted image, the explanation can change drastically, and that opens the door for exploiting the similarity between superpixels regardless of the distortion. Based on this observation, we propose to weight the samples used to train the local surrogate model with perceptual distances. Perceptual distances aim to model how humans perceive distortions in images based off human visual experiments. Although the explicit connection in between perceptual metrics and statistical learning has only recently been shown~\cite{hepburn2021relation}, perceptual considerations have been introduced before in different stages of the learning process, including in deep network architecture design~\cite{perceptnet2020} or shaping the objective function for deep models~\cite{laplacian2019}. However, in including this information in the training of the local surrogate model, we are asking the explainer to find the most informative features, regardless of the distortion applied to the image.

The paper is structured as follows. Sec.~\ref{sec:surrogates} describes the procedure for generating explanations using surrogate explainers in the setting of image classification and a proposed explanation distance that is independent of interpretable domains from which samples are generated. Sec.~\ref{sec:experiments} presents the empirical evaluation of the effect of perceptual and non-perceptual distances in achieving robust explanations in image classification tasks under distortions.

\section{Robust Surrogate Explainers}\label{sec:surrogates}
First introduced as local interpretable model-agnostic explanations (LIME)~\cite{ribeiro2016should}, surrogate explainers attempt to find a simpler, usually linear, model that is accurate on the decision boundary close to a sample data point $x$.
We define an explanation as 
\begin{equation}
    \text{exp}_{x} = \argmin_{g\in \mathcal{G}} \mathcal{L}(f, g, \pi_x) + \Omega(g),
\end{equation}
where $\mathcal{L}(f, g, \pi_x)$ defines the fit of surrogate model $g$ from model family $\mathcal{G}$ to the black-box model $f$ in the neighbourhood $\pi_x$ of a query data point $x$ that belongs to the same distribution $\mathcal{X}$ used to train the black-box model. $\Omega(g)$ is a penalisation on the complexity of model $g$. Intuitively this formulation is attempting to find the surrogate model $g$ that best fits the black-box model $f$ only around the neighbourhood $\pi$. $g$ is usually trained on the neighbourhood in an interpretable feature domain, usually a binary vector encoding the presence of human-understandable features in the data.

\subsection{Building surrogate explainers}
Constructing surrogate explainers can be decomposed into three main stages; interpretable data representation, data sampling and explanation generation~\cite{sokol2019blimey}.

\paragraph*{Data representation} The interpretable data representation is a transformation from the data domain $\mathcal{X}$ to an interpretable domain $\mathcal{Z}$. For images, superpixels are found using a segmentation algorithm defined by the user. The interpretable representation is then a binary vector encoding the state of a superpixel within the image; whether it has been ablated or not.

\paragraph*{Data sampling} The data sampling step defines the points that create the neighbourhood $\pi_x$ around the query data point $x$. Sampling is usually performed in the interpretable domain and in order to train the surrogate model $g$, the outputs of the black-box model that correspond to data points in the neighbourhood $f(\pi_x)$ are required. If the data was sampled in the interpretable domain, it needs to be transformed back into the original data domain $\mathcal{X}$. For images, this stage is done by sampling binary vectors $z'$ from a discrete uniform distribution and creating images $z$ in the original domain with superpixels ablated according to the binary feature defined in the sampled vectors. The ablation can mean either setting all pixel values to $0$ if they are within the ablated superpixel, or setting the pixels to the mean value of the superpixel. These images can then be used as an input in the model $f$ to get the outputs. We then define the neighbourhood as
\begin{equation}\label{eq:pi}
    \pi_x(x, z) = \exp\left(\frac{-D(x, z)^2}{\sigma^2}\right)
\end{equation}
where $D$ is a distance, which can either be in $\mathcal{X}$ or $\mathcal{Z}$ and $\sigma$ is the width of the exponential kernel. By default, the standard surrogate implementations use the cosine distance in the binary vector representation $\mathcal{Z}$, between a binary vector of all ones representing the original image $x'$ and samples $z'$.

\paragraph*{Explanation generation} The final stage is training a interpretable model $g$, usually a simple linear model, on the sampled data. This model aims to predict the outputs of the black-box model for the sampled neighbourhood $f(z)$. The regression targets used are the prediction probabilities from the black-box model for the explanation class(es), defining a locally weighted square loss
\begin{equation}
    \mathcal{L}(f, g, \pi_x) = \sum_{z, z' \in \mathcal{Z}} \pi_x(x, z) (f(z) - g(z'))^2.
\end{equation}
$\Omega(g)$ introduces regularisation on the model $g$. For example, if $\Omega(g)$ is the L2 norm of the weights of $g$, the local linear model becomes ridge regression.

\subsection{Surrogate explainers in the wild}
When image models are deployed in the real-world, distorted images become more frequent~\cite{hendrycks2019benchmarking}. A user taking an out of focus image is more likely than finding an out of focus image in the training dataset of the model. Of course, this data shift can cause the models predictions to change due to the model not being exposed to distorted images during training. However, even if the predictions of the model remain constant, the ability to generate useful explanations is not guaranteed. This is due to the construction of the interpretable domain, in which the generation of image explanations involves  image segmentation. For example, if the distortion applied is Gaussian noise, then the lines within the image may be blurred leading to a different segmentation result.

The segmentation method that is used to generate the interpretable data domain can be severely affected by distortions, causing an image and its distorted counterpart to share the same prediction but have very different explanations. This can either be addressed in making the segmentation method more robust or making the surrogate model $g$ more robust to distortions. In this paper, we focus on the latter.

We propose the use of metrics that take into account distortions in images as distance $D$ in Eq.~\ref{eq:pi}, attempting to capture the similarity in images irrespective of the distortion applied to them. In order to do so, $D$ can be a perceptual metrics, measuring the perceptual distance between the original image and the sampled points in domain $\mathcal{X}$, namely $D(x, z)$.

\subsection{Explanation distance}
Evaluating explanations and understanding the information they convey are ongoing and complex problems~\cite{poyiadzi2021overlooked,sokol2020one}. Instead of judging whether one explanation is better or worse, we simply find a distance between explanations that can be used irrespective of the interpretable data domain of the explanations.

Due to the distortions applied to the image, the segmentation and therefore the interpretable data domain can be different and taking a distance between interpretable domains is not possible. Therefore we project the explanation back to the image domain and measure a distance there. For each image, we construct a matrix $\mathbf{E_k}$ where each entry $E_k{(i,j)}$ is the importance value value in the explanation for class $k$ of the superpixel where pixel $(i,j)$ belongs. The distance used is then the average sum of squared error between the matrix constructed for the reference and distorted image explanations for all the explained classes $K$,
\begin{equation}
    d_{exp} = \frac{1}{K} \sum_{k=0}^K |\mathbf{E_k} - \mathbf{\tilde{E}_k}|_F^2.
\end{equation}

\subsection{Perceptual Distances}\label{sec:perceptual_distances}
In order to capture the human visual systems ability to perceive changes across a set of images, practitioners have proposed models that attempt to recreate certain psychophysical phenomena observed in humans. 

One of the main perceptual metrics used in practice is structural similarity and its multi-scale variant, \textit{multi-scale structural similarity} (MS-SSIM)~\cite{wang2003multiscale}, which aims to measure the distance between statistics of the reference and distorted images at various scales. This distance is based on the principle that perceptual structural similarity of the image will be preserved despite the distortion. Differently, the \textit{normalised Laplacian pyramid distance} (NLPD)~\cite{laparra2016,laparra2017perceptually} uses a Laplacian pyramid with local normalisation at the output of each stage. The image is encoded by performing convolutions with a low-pass filter and subtracting this from the original image. This is then repeated for as many stages as there are in the pyramid. The output of each stage is then locally normalised using divisive normalisation. After the transformation is applied to the reference and distorted image, a distance can be taken this in space. This distance differs from MS-SSIM as it is based on the visibility of errors~\cite{Watson93}, and acts as a transformation to a more perceptually meaningful domain. Once transformed, simple distances reflect the perceptual similarity between two images. 
Both these distances have been shown to correlate well with human perception -- with NLPD performing the best. In the experiments we will use both MS-SSIM and NLPD as our perceptual distances as to cover both principles; structural similarity and visibility of errors.

\begin{figure*}[tb]
    \centering
    \includegraphics[width=\textwidth]{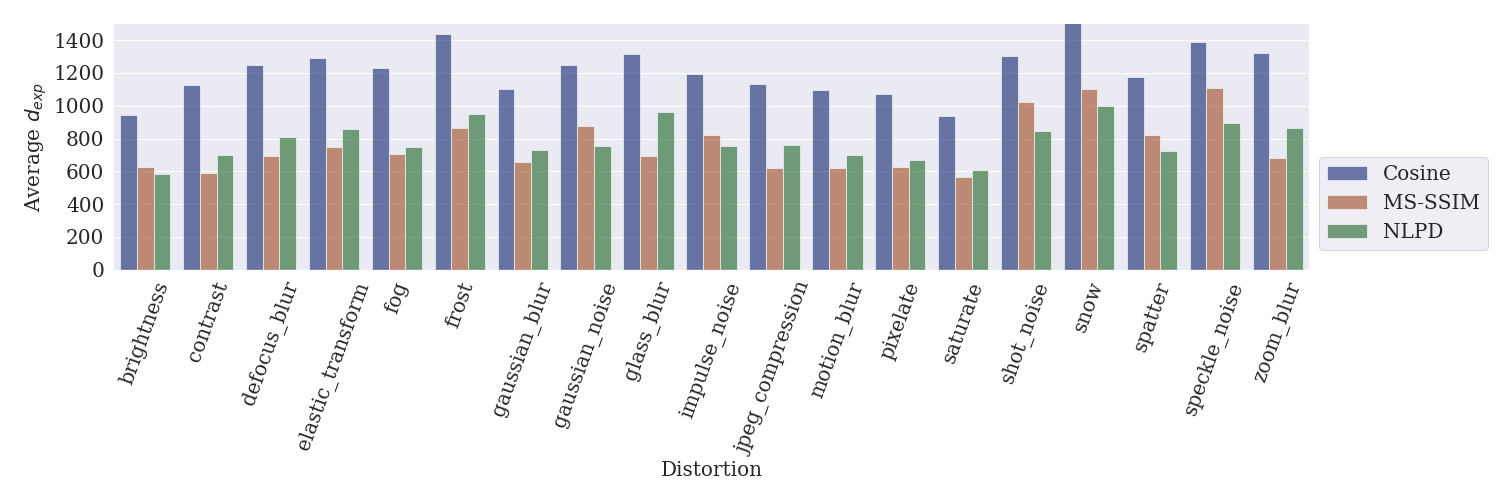}
    \caption{Average $d_{exp}$ for $K=2$ of a number of distortions for explanations generated using 3 different kernalised distances; cosine, MS-SSIM and NLPD. For a further breakdown of the pairs of images used to calculate these averages, see Table (\ref{tab:dataset})}
    \label{fig:results}
\end{figure*}

\section{Experiments}\label{sec:experiments}
In order to evaluate the effect of using perceptual metrics to weight the distances in the local neighbourhood $\pi_x$, we train local surrogate explainers on a reference image and an image with distortions applied to it. 

\paragraph*{Dataset} The reference images come from the Imagenet validation dataset~\cite{deng2009imagenet} while the distorted images are taken from the Imagenet-C dataset~\cite{hendrycks2019benchmarking}. The Imagenet-C dataset contains distorted versions of the Imagenet validation set, with 15 corruptions at 5 increasing severities. It covers natural distortions, like weather effects and artificial distortions like Gaussian blurring. For a full description of the dataset  see \cite{deng2009imagenet}.

\begin{table}[h]
\centering
\begin{tabular}{|l|llllll|}
\hline
\multicolumn{1}{|c|}{\multirow{2}{*}{Distortion}} & \multicolumn{6}{c|}{Strength}  \\ \cline{2-7} 
\multicolumn{1}{|c|}{}                            & 1  & 2  & 3  & 4  & 5  & Total \\ \hline
Brightness                                        & 50 & 45 & 40 & 37 & 27 & 199   \\
Contrast                                          & 44 & 41 & 35 & 21 & 7  & 148   \\
Defocus Blur                                      & 29 & 25 & 14 & 9  & 6  & 83    \\
Elastic                                           & 34 & 16 & 28 & 19 & 8  & 105   \\
Fog                                               & 25 & 26 & 21 & 19 & 11 & 102   \\
Frost                                             & 34 & 23 & 14 & 13 & 13 & 97    \\
Gaussian Blur                                     & 34 & 25 & 15 & 7  & 4  & 85    \\
Gaussian Noise                                    & 37 & 29 & 21 & 14 & 8  & 109   \\
Glass Blur                                        & 29 & 17 & 5  & 4  & 3  & 58    \\
Impulse Noise                                     & 25 & 18 & 18 & 10 & 4  & 75    \\
JPEG Compression                                  & 36 & 37 & 38 & 28 & 22 & 161   \\
Motion Blur                                         & 35 & 20 & 14 & 7  & 4  & 80    \\
Pixelate                                          & 37 & 42 & 31 & 31 & 20 & 161   \\
Saturate                                          & 45 & 39 & 48 & 37 & 30 & 199   \\
Shot Noise                                           & 32 & 25 & 20 & 11 & 9  & 97    \\
Snow                                              & 32 & 19 & 12 & 14 & 12 & 89    \\
Spatter                                           & 44 & 39 & 28 & 19 & 13 & 143   \\
Speckle Noise                                        & 35 & 25 & 21 & 16 & 11 & 108   \\
Zoom Blur                                           & 28 & 17 & 12 & 9  & 6  & 72    \\ \hline
\end{tabular}
\caption{Breakdown of images used where the distorted and reference image have the same top-2 classes predicted.} 
\label{tab:dataset}
\end{table}

\paragraph*{Experimental framework} In order to compare the generated explanations, their objective has to be the same -- to explain class $k$. When generating explanations for a reference and distorted image, we need to ensure that the model assigns similar predictions to both images. We only use images where the top-$K$ classes predicted by the model are shared. Due to strong distortions changing the predictions of the model, there is only a subset of the distorted images from Imagenet-C that share the top-$K$ predicted classes. We randomly sampled 70 reference images from the validation set of Imagenet and the corresponding distorted images from Imagenet-C. We only use 70 images as it is computationally expensive to train a surrogate explainer for each image. We took distorted images that share the top-2 predicted classes, and generated explanations for these classes. The number of images found with shared top-2 predicted classes for each distortion is reported in Table (\ref{tab:dataset}). The classification model that we are generating explanations for is Inception-V3~\cite{szegedy2016rethinking}. For each class, we generated 3 explanations, each with a different weighted kernalised distance $D$. The three distances used were: cosine similarity between the binary interpretable representations $x'$ and $z'$, MS-SSIM and NLPD both in the image domain $x$ and $z$. The explanation generation procedure used is the same as in \cite{ribeiro2016should}; a superpixel interpretable domain; $1000$ points uniformly sampled in binary space and ridge regression as a surrogate model with the samples weighted by one of the 3 distances. The exponential kernel with width $0.25$ is used. The explanation distance $d_{exp}$ is computed between the explanations generated for each reference-distorted image pair and each distance. For all experiments we used the FAT-Forensics package~\cite{sokol2020fat-forensics}.

\paragraph*{Results}
Overall the average $d_{exp}$ for all distortions and strengths is significantly decreased when using perceptual metrics. For cosine similarity, the overall average $d_{exp}$ is $1181\pm 1012$, for MS-SSIM $739 \pm 745$ and for NLPD $756 \pm 704$. A further breakdown can be seen in Fig.~(\ref{fig:results}). For all distortions, the perceptual metrics outperform the cosine similarity. For distortions that act in a locally, e.g. shot noise, impulse noise and spatter, NLPD performs the best and for distortions involving a smoothing, e.g. zoom blur, glass blur and defocus blur, MS-SSIM is the best performing distance. This reflects the properties of the perceptual metrics, with the divisive normalisation in NLPD penalising sudden local changes whilst MS-SSIM computes statistics of the images over a window and is able to notice that within a window, gradual changes are applied. The standard deviations of the average $d_{exp}$ are high due to the fact that different distortions applied to different images can alter the content of the image differently. Objects with a rigid structure like a fence would be segmented very differently when applying an elastic transform as straight lines will have been distorted. However, take an image of the sea and apply the same distortion and the image will be perceptually similar to the original.

\section{Conclusion}\label{sec:conclusion}
We introduced a methodology to evaluate the robustness of surrogate-based explanations in the presence of distortions for image classification. We showed empirically the classification output can be similar yet the resulting explanation can differ for a reference image and distorted image. To address this, we tested two perceptual metrics in the training of the local surrogate explainer, and empirically showed that these create more robust and coherent explanations when images are subject to distortions.


\clearpage
\bibliographystyle{IEEEbib}
\bibliography{ref}

\end{document}